\documentclass[runningheads]{llncs}
\usepackage[T1]{fontenc}
\usepackage{graphicx,verbatim}
\usepackage{xcolor}
\usepackage{bm}
\usepackage{multirow}
\usepackage{multicol}
\usepackage{amsmath}
\usepackage{amssymb}
\usepackage[colorlinks]{hyperref}
\usepackage{cite}

\begin{document}
\title{Sparse Reconstruction of Optical Doppler Tomography with Alternative State Space Model and Attention}
%
\author{Zhenghong Li \and
Jiaxiang Ren \and
Wensheng Cheng \and 
Yanzuo Liu \and
Congwu Du \and
Yingtian Pan \and 
Haibin Ling}
\authorrunning{Z. Li et al.}
%
\institute{Stony Brook University\\
\email{\{zhenghli, hling\}@cs.stonybrook.edu}}


\maketitle              
\begin{abstract}
Optical coherence Doppler tomography (ODT) is an emerging blood flow imaging technique. The fundamental unit of ODT is the 1D depth-resolved trace named raw A-scans (or A-line). A 2D ODT image (B-scan) is formed by reconstructing a cross-sectional flow image via Doppler phase-subtraction of raw A-scans along B-line. To obtain a high-fidelity B-scan, densely sampled A-scans are required currently, leading to prolonged scanning time and increased storage demands. Addressing this issue, we propose a novel sparse ODT reconstruction framework with an Alternative State Space Attention Network (ASSAN) that effectively reduces raw A-scans needed. Inspired by the distinct distributions of information along A-line and B-line, ASSAN applies 1D State Space Model (SSM) to each A-line to learn the intra-A-scan representation, while using 1D gated self-attention along B-line to capture the inter-A-scan features. In addition, an effective feedforward network based on sequential 1D convolutions along different axes is employed to enhance the local feature. In validation experiments on real animal data, ASSAN shows clear effectiveness in the reconstruction in comparison with state-of-the-art reconstruction methods. The source code is available at: \href{https://github.com/ZhenghLi/ASSAN}{https://github.com/ZhenghLi/ASSAN}.

\keywords{Optical Doppler Tomography \and Sparse Reconstruction.}

\end{abstract}

\section{Introduction}
\label{sec:intro}
Optical Doppler Tomography (ODT), \textit{a.k.a.} Doppler Optical Coherence Tomography, is an emerging technique for blood flow visualization and quantification~\cite{leitgeb2014doppler,zhao2000phase,chen2008doppler}. It acquires high-resolution and high-contrast tomographic images~\cite{li2020deep} and has many potential clinical applications such as vascular disease monitoring~\cite{leitgeb2014doppler}. To detect the vessels with slow blood flows, dense scanning is usually employed~\cite{wang2007three} in ODT since the sensitivity to the motion depends on the sampling time~\cite{spaide2018optical}. Therefore, the traditional pipeline often suffers from long scanning time and serious storage burden. For example, for a 3D ODT volume of 500 2D B-scans, each raw B-scan takes nearly 4s and 55MB in our experiments.

\begin{figure*}[!t]
\centering
\includegraphics[width=1\linewidth]{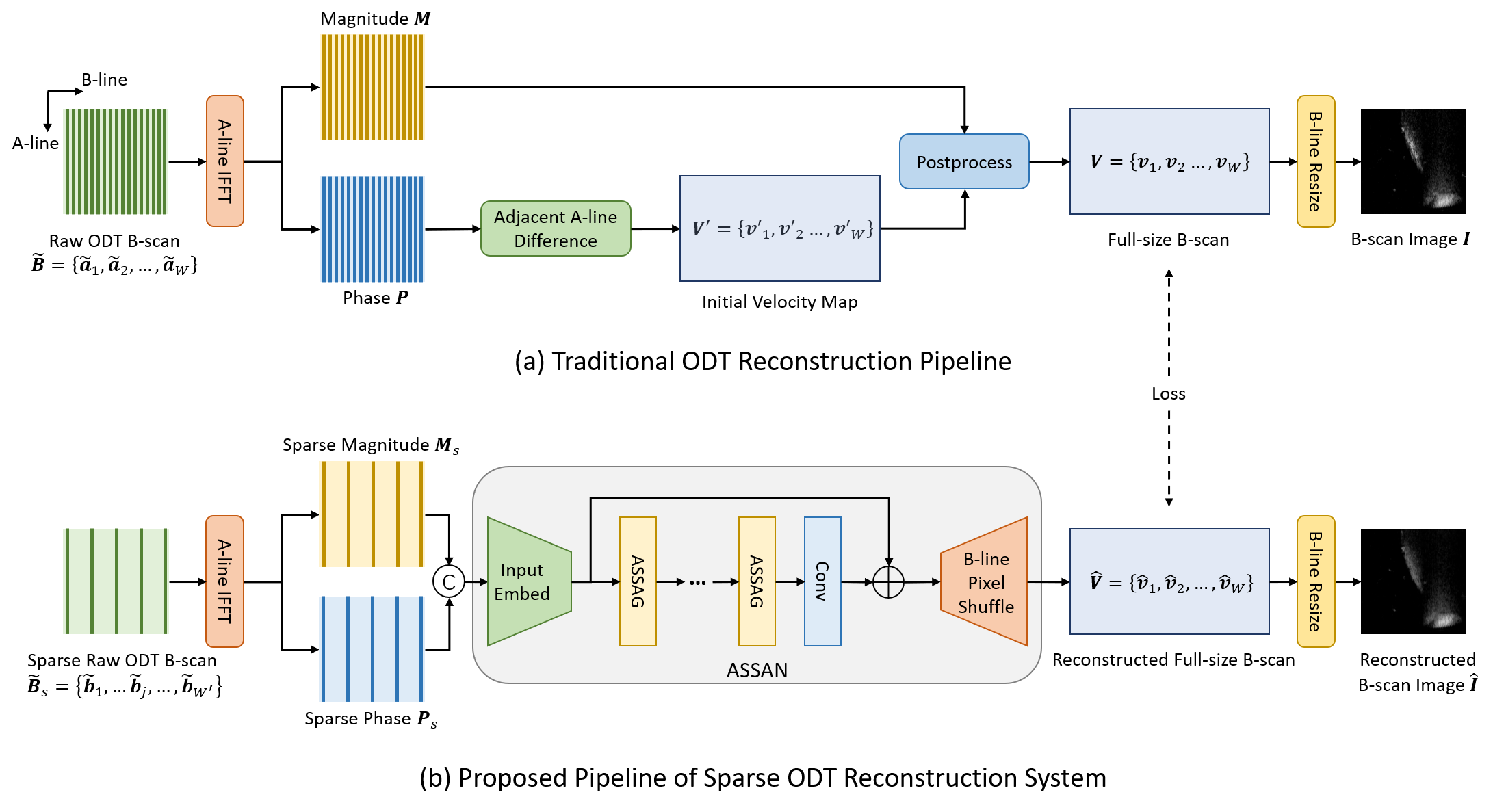}
    \caption{ODT reconstruction. For sparse reconstruction (b), the sparse raw B-scan $\Tilde{\bm{B}}_s$ is downsampled from $\Tilde{\bm{B}}$ with a stride $\delta$. $\Tilde{\bm{B}}_s$ is forwarded to our Alternative State Space Attention Network (ASSAN) to reconstruct the B-scan image (details in Sec.~\ref{sec:formulation}).}
\label{fig:odteg}
\end{figure*}

In a typical ODT image reconstruction pipeline (Fig.~\ref{fig:odteg}(a)), the basic scan of raw ODT signals is along the A-line (depth) called raw A-scan encoding the depth information. Sliding the scanning sensor along the B-line (width) obtains a 2D signal called raw B-scan. Then, the final B-scan is achieved after a series of magnitude-phase processing as an indication of blood flow. 

To solve the dense scanning issue, we propose, for the first time, a deep learning-based sparse ODT reconstruction pipeline (Fig.~\ref{fig:odteg}(b)). The new pipeline requires much fewer A-scans, thus reducing the scanning time and storage burdens. There have been many works on sparse reconstruction based on compressive sensing techniques~\cite{zhang2015survey,marques2018review}, which are applied in various domains, such as magnetic resonance imaging~\cite{ning2015sparse,yang2016sparse}, radar~\cite{potter2008sparse,wei2010sparse}, and OCT~\cite{fang2016segmentation}. However, adapting these techniques to sparse ODT reconstruction is non-trivial and unexplored. Besides, simply applying generic image feature extraction methods for this task may only get suboptimal performance, due to the fundamental difference between natural images and ODT B-scans: information distribution along A-line differs from that along B-line. The A-line is depth-correlated and thus heterogeneous, while the B-line is nearly homogeneous. Besides, the sparse sampling is constrained to the B-line, and no downsampling happens to A-scans. Particularly, since A-line is long, to handle the full-depth information, we propose to use Mamba~\cite{gu2023mamba}, a recently introduced State Space Model (SSM)~\cite{kalman1960new,gu2021efficiently,gu2021combining}, which can capture long-range dependency with linear complexity.

Motivated by the above analysis, we develop a novel Alternative State Space Attention Network (ASSAN) (Figure \ref{fig:odteg}(b)) for sparse ODT reconstruction, with five main components: (1) an input block for initial encoding, (2) A-line state space blocks to handle the depth-correlated A-line information, (3) novel B-line gated attention blocks to model the inter-A-scan features along each B-line, (4) an enhancement feedforward network to learn local features, and (5) a B-line pixel shuffle layer to upscale the B-line for reconstruction. 

With all these components, ASSAN effectively captures the unique characteristics of ODT B-scans for accurate reconstruction. To our best knowledge, this is the first attempt for sparse ODT reconstruction and the first report of combining Mamba and self-attention modules to image reconstruction tasks. Experiments demonstrate that our method has exhibited promising results in comparison with state-of-the-arts image reconstruction methods. Besides, the proposed framework is expected to significantly save scanning time (critical to studying flow dynamics to various brain stimulation) and memory while keeping good results.
To summarize, our main contributions include: 
\begin{enumerate}
    \item the first sparse reconstruction pipeline for ODT image generation to save the scanning time and storage,
    \item a novel alternative SSM attention model along different axes for image reconstruction,
    \item an effective gated self-attention model to refine the modeling of inter-A-scan features along the B-line, and
    \item the effective sequential Conv1D-based FFN to enhance local features.
\end{enumerate}

\section{Alternative State Space Attention Network}
\subsection{Problem Formulation}
\label{sec:formulation}
Our task is to reconstruct an ODT B-scan image $\hat{\bm{I}}$ from the sparse raw B-scans $\Tilde{\bm{B}}_s$ to speed up scanning and save storage. The traditional ODT reconstruction pipeline with dense scanning is shown in Fig.~\ref{fig:odteg}(a). The basic scan of ODT is called raw A-scan $\Tilde{\bm{a}}_i$, encoding the depth (A-line) information in the frequency domain. A 2D raw signal named raw B-scan $\Tilde{\bm{B}} = \{\Tilde{\bm{a}}_i\}_{i=1}^W$, containing $W$ raw A-scans, is acquired by sliding the OCT probing beam along the B-line (width). 1D Inverse Fast Fourier Transform (IFFT) is applied to each raw A-scan to decode the magnitude and phase along the depth. The 1D phase difference of successive A-scans represents the Doppler frequency shift~\cite{zhao2000phase}, estimating the initial blood flow velocity map $\bm{V}'$. The magnitude is usually used as a mask for denoising in post-process to get the reconstructed full-size B-scan $\bm{V} = \{\bm{v}_i\}_{i=1}^W$. Due to dense sampling, the aspect ratio of $\bm{V}$ is often too small. Finally, a B-line resize operation is used to shrink the width of $\bm{V}$ to obtain the final B-scan image $\bm{I}$. 

To solve the dense scanning issue of the traditional pipeline, we propose a deep learning-based sparse reconstruction pipeline for ODT imaging using fewer raw A-scans, as illustrated in Fig.~\ref{fig:odteg}(b). The goal is to reconstruct a high-quality ODT image $\hat{\bm{I}}$ from sparsely sampled raw A-scans stored in $\Tilde{\bm{B}}_s = \{\Tilde{\bm{b}}_j= \Tilde{\bm{a}}_{1+(j-1)\delta}\}_{j=1}^{W'}$, where $\delta$ is sampling stride along B-line, $W' = W /\delta$ is the number of sparsely sampled raw A-scans. To maximize the usage of available data for training, we let the network predict the full-size B-scan $\hat{\bm{V}} = \{\hat{\bm{v}}_i\}_{i=1}^{W}$ and then resize the B-line of $\hat{\bm{V}}$ as the post-process to get the final result $\hat{\bm{I}}$.

\subsection{Overall Pipeline of ASSAN}
The overall pipeline of our method is presented in Fig.~\ref{fig:odteg}(b). The input to the whole system is a sparsely sampled raw ODT B-scan $\Tilde{\bm{B}}_s$. Following the traditional pre-process, 1D IFFT is applied to each raw A-scan $\Tilde{\bm{b}}_j$ in $\Tilde{\bm{B}}_s$ to transform the information from the frequency domain representation to the spatial magnitude $\bm{M}_s$ and phase $\bm{P}_s$ responses along the depth. To encourage rich interaction of the complementary information in magnitude and phase, we concatenate them along the channel dimension as the input to the ASSAN (denoted as $\bm{X}_{in}$). The network input is first encoded via a convolution layer and then processed by a sequence of $G$ Alternative State Space Attention Groups (ASSAG). To maximize the utilization of the available supervision data, a B-line Pixel Shuffle (B-PS) layer is employed to upscale the B-line {by rearranging the channel and width dimension} to predict the full-size B-scan $\hat{\bm{V}}$. The network can be summarized as:
\begin{equation}
    \begin{split}
        \bm{X}_0 = \text{Conv}(\bm{X}_{in})&,
        \bm{X}_g = \text{ASSAG}_g (\bm{X}_{g-1}),~~g = 1, 2, \ldots, G \\
        \hat{\bm{V}} &= \text{B-PS} (\bm{X}_0 + \text{Conv}(\bm{X}_G)).
    \end{split}
    \label{eq:pipe}
\end{equation}
We employ the pixel $L_2$ loss as the optimization function. Finally, the B-line of the predicted $\hat{\bm{V}}$ is resized to get the final B-scan image $\hat{\bm{I}}$ for visualization.

\begin{figure*}[!t]
\centering
\includegraphics[width=1.0\linewidth]{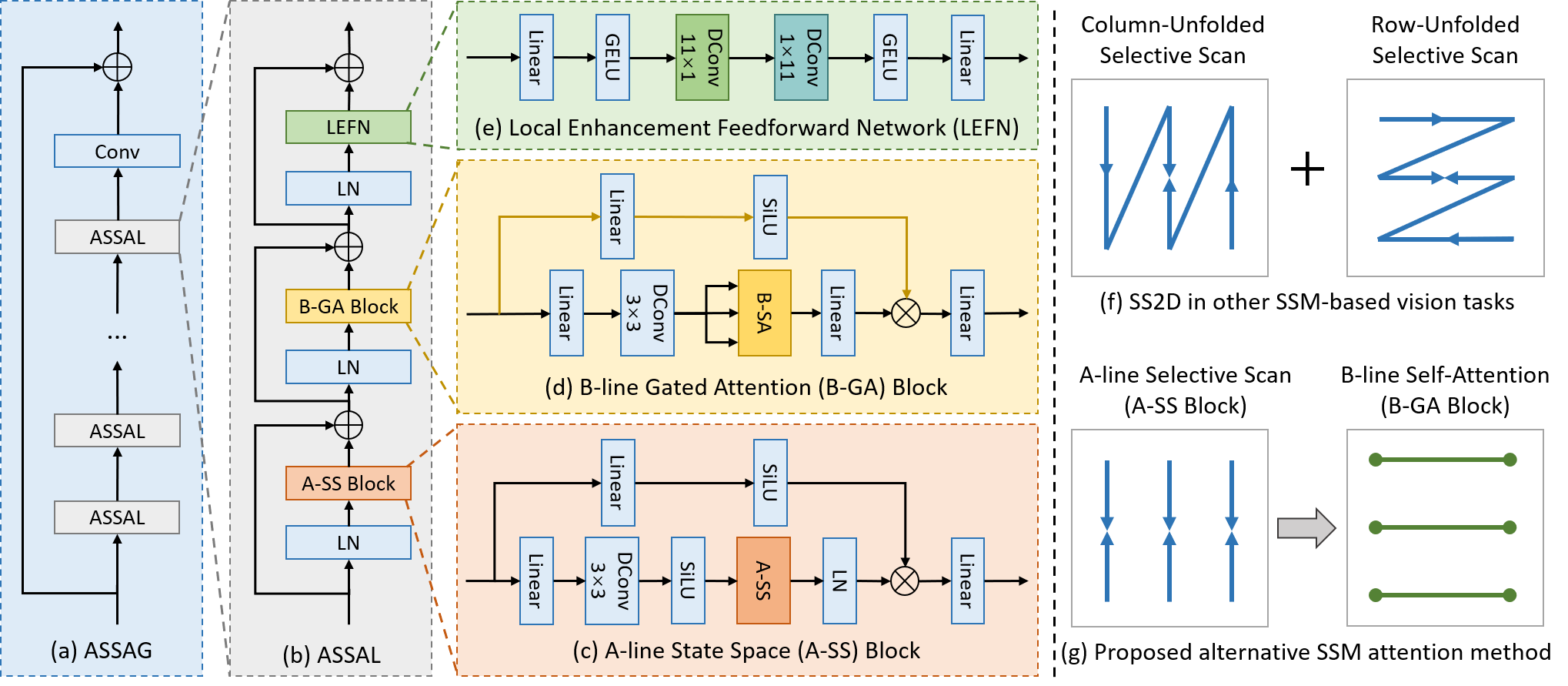}
    \caption{Network blocks and SSM-based scanning strategies. Blocks: (a) Alternative State Space Attention Group (ASSAG), (b) Alternative State Space Attention Layer (ASSAL), (c) A-line State Space (A-SS) Block, (d) B-line Gated Attention (B-GA) Block, and (e) Local Enhancement Feedforward Network (LEFN). Scans: (f) Cross-scan 2D Selective Scan methods~\cite{liu2024vmamba,guo2024mambair} unfold the image to sequences along rows and columns, and scan them via SSM in parallel. (g) Our method first bidirectionally scans each A-line (column) via SSM, and then applies self-attention to each B-line (row).}
\label{fig:blockscan}
\end{figure*}

\subsection{Alternative State Space Attention Group}
The Alternative State Space Attention Group (ASSAG), shown in Fig.~\ref{fig:blockscan}(a), consists of a sequence of Alternative State Space Attention Layers (ASSAL) and a convolution layer for deep feature extraction. 
The ASSAL's architecture and feature processing order are illustrated in Fig.~\ref{fig:blockscan}(b) and Fig.~\ref{fig:blockscan}(g). To model the different distributions along the A-line and B-line, we alternatively apply the A-line State Space (A-SS) Block and the B-line Gated Attention (B-GA) Block to model the A-line and B-line respectively. Then we employ a Local Enhancement Feedforward Network (LEFN) to enhance local features via Conv1Ds along different axes. LayerNorm (LN) is employed for normalization.

\noindent\textbf{A-line State Space Block (Fig.~\ref{fig:blockscan}(c)).}
Since an ODT A-scan is correlated with the depth, and the signals of large and/or vertically growing vessels may span long, we employ Mamba~\cite{gu2023mamba} to model full-depth information for its strong sequence modeling capability with linear complexity. In contrast, as widely applied self-attention's complexity is quadratic to the sequence length and the depth of an A-scan is typically much longer than a usual window attention~\cite{liang2021swinir} (e.g., 512 vs 64), self-attention is not suitable to model full depth A-scans. 

Unlike most visual Mamba approaches~\cite{liu2024vmamba,guo2024mambair}, which directly model the entire image via 2D Selective Scan (SS2D) (Fig.~\ref{fig:blockscan}(f)), we introduce a novel A-line Selective Scan (A-SS) mechanism tailored to A-scans (Fig.~\ref{fig:blockscan}(g) left). Specifically, A-SS conducts 1D bidirectional scans on each A-line in parallel using Mamba's 1D selective scan. Compared to SS2D, which unfolds feature maps into sequences by concatenating rows or columns—potentially disrupting depth information—A-SS more effectively models the depth-correlated intra-A-scan features. Given an input $\bm{X}$, an A-SS Block can be represented as:
\begin{equation}
    \begin{split}
        \bm{X}_1 = \text{SiLU}(\text{Linear}(\bm{X}))&, 
        \bm{X}_2 = \text{LN}\big(\text{A-SS}\big( \text{SiLU}(\text{DConv}(\text{Linear}(\bm{X}))) \big) \big), \\
        \bm{X}_\text{out} &= \text{Linear}(\bm{X}_1 \odot \bm{X}_2).
    \end{split}
    \label{eq:assb}
\end{equation}
where SiLU~\cite{elfwing2018sigmoid} is the Sigmoid-weighted Linear Unit activation function, DConv denotes the depth-wise convolution, and $\odot$ denotes the element-wise product.

\noindent\textbf{B-line Gated Attention Block (Fig.~\ref{fig:blockscan}(d)).}
We first introduce a B-line Self-Attention (B-SA) module to model inter-A-scan features, which processes each B-line via 1D self-attention~\cite{vaswani2017attention} in parallel (Fig.~\ref{fig:blockscan}(g) right). Since vessels can span long along B-line, capturing long-range information is desired for inter-A-scan modeling. Unlike the depth-correlated A-line, B-line information is nearly homogeneous, making self-attention well-suited. 
Besides, as sequentially applying bidirectional 1D Mamba across different axes severely affects the throughput~\cite{liu2024vmamba}, using B-SA instead of Mamba also enhances runtime. 

However, simply applying the B-SA may not yield optimal performance, as it ignores the distinct behaviors of deep features along this axis. Therefore, uniformly processing B-line features that encode both magnitude and phase information is suboptimal. To further boost the B-line modeling, in addition to the B-SA, we propose a gated attention mechanism to account for the distinct properties of magnitude and phase along the B-line. This mechanism addresses the issue by processing the features in two branches. In one branch, B-SA is employed to model the inter-A-scan features along the B-line, similar to phase processing in the traditional pipeline. In the other branch, the input feature map is transformed into a score mask to refine the features learned by B-SA, analogous to the traditional processing of magnitude. {Although there have been some works on gated attention~\cite{yang2024gated,han2025demystify}, none of them explores the application to low-level vision tasks}. Given an input $\bm{X}$, a B-GA Block can be represented as:
\begin{equation}
    \begin{split}
        \bm{X}_1 = \text{SiLU}(\text{Linear}(\bm{X}))&, 
        \bm{X}_2 = \text{Linear}\big(\text{B-SA}\big(\text{DConv}(\text{Linear}(\bm{X})) \big) \big), \\
        \bm{X}_\text{out} &= \text{Linear}(\bm{X}_1 \odot \bm{X}_2).
    \end{split}
    \label{eq:bgab}
\end{equation}

\noindent\textbf{Local Enhancement Feedforward Network (Fig.~\ref{fig:blockscan}(e)).}
To further enhance the local information, we introduce the Local Enhancement Feedforward Network (LEFN) by inserting Conv1Ds along different axes into a multi-layer perception (MLP) with GELU~\cite{hendrycks2016gaussian} activations. Because the above A-SS Block and B-GA Block primarily focus on long-range features, a convolution-based FFN with a large local receptive field could help enhance the local features. Inspired by~\cite{szegedy2016rethinking}, instead of directly using a large Conv2D kernel, we factorize the Conv2D kernel into two Conv1D kernels along different axes. This way, LEFN not only follows the separate processing of different axes but also brings a large receptive field with fewer parameters.

\section{Experiments}
\label{sec:experiment}
\subsection{Datasets and Implementation Details}
\noindent\textbf{Datasets.} The datasets in previous related works~\cite{li2020deep,jiang2020comparative} contain only 6 volumes in total. 
Therefore, we collected two datasets, Mouse Cortex Dataset 2021 (MCD-21) and Mouse Cortex Dataset 2023 (MCD-23), each gathered by different experts on different groups of mice. MCD-21 includes 16 volumes (10 for training and 6 for testing) from awake and anesthetized mice. MCD-23 consists of 13 volumes (8 for training and 5 for testing) from anesthetized mice. The train and test volumes are from different mice to better test the generalization capability of the models. Each volume includes 500 densely scanned raw B-scans, and each raw B-scan consists of 12600 raw A-scans. The depth of the IFFT-related pre-processed input to the networks is 512. For practical evaluation, we calculate the PSNR and SSIM for each final B-scan image and crop out deep background regions in the test images to minimize the influence of background noise. 

\noindent\textbf{Implementation details.}
We conduct experiments with the sparse sampling factors 2 and 4 (denoted as $\times 2$ and $\times 4$). In our ASSAN, we employ 4 ASSAGs, each with 6 ASSALs. The feature embedding dimension is set to 60. The size of a pre-processed sparse B-scan patch is $512\times64$. The batch size is set to 8. We employ Adam as the optimizer with $\beta_1 = 0.9$, $\beta_2 = 0.99$. The number of training iterations is set to 200K. The learning rate scheduler is CosineAnnealingLR~\cite{loshchilov2016sgdr} whose initial and minimum learning rates are $2e-4$ and $1e-6$.

{Since there is no previous work for sparse ODT reconstruction, we compare ASSAN with single-image super-resolution methods~\cite{lim2017enhanced,zhang2018image,dai2019second,niu2020single,chen2021pre,liang2021swinir,chen2023activating,li2023efficient,zhou2023srformer,chen2023dual,ray2024cfat,zhang2024transcending,guo2024mambair} as they also upsample the inputs. We compare with five transformer-based state-of-the-arts: 
SwinIR~\cite{liang2021swinir}, HAT~\cite{chen2023activating}, SRFormer~\cite{zhou2023srformer}, DAT~\cite{chen2023dual}, CFAT~\cite{ray2024cfat}, and one SSM-based: MambaIR~\cite{guo2024mambair}. We apply the same input embedding and replace their upsampling layer~\cite{shi2016real} with our B-PS layer to make them applicable.

\subsection{Experimental Results}
\textbf{Quantitative results.}
Table~\ref{table:result} summarizes the experimental results. Most of the transformer-based methods do not perform well, with the exception of DAT, which also uses channel-wise self-attention to capture global information. However, because channel-wise self-attention is not directly based on spatial features, it lacks accuracy. Though MambaIR outperforms transformer-based methods, its results are still suboptimal due to the SS2D scanning method, which can confuse A-line and B-line information. In contrast, our ASSAN separates A-line and B-line processing with alternative SSM and attention along different axes, thus achieving superior performance compared to other methods.

\begin{figure*}[!t]
\centering
\includegraphics[width=1\linewidth]{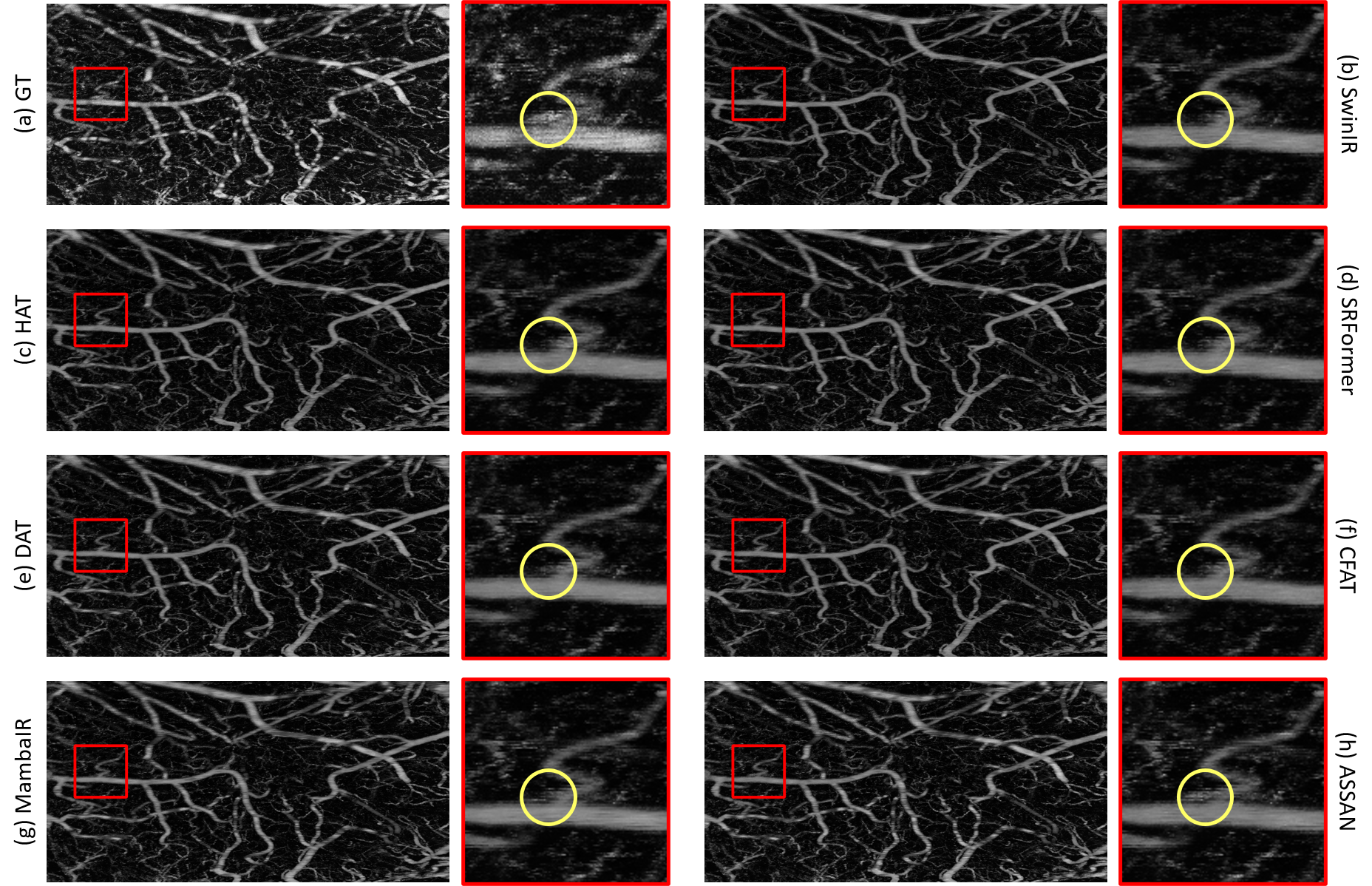}
    \caption{Qualitative results of $\times4$ sparsity on MCD-23. Regions of interest are zoomed in. Challenging cases are marked by circles.}
\label{fig:result}
\end{figure*}

\begin{table}[!tbp]
\caption{Comparison of sparse ODT reconstruction results.}
\begin{center}
\small
\begin{tabular}{l|cccc|cccc}
\hline
\multirow{3}{*}{Method} & \multicolumn{4}{c|}{x2 Sparsity}                                                         & \multicolumn{4}{c}{x4 Sparsity}                                                          \\ \cline{2-9} 
                        & \multicolumn{2}{c|}{MCD-21}                           & \multicolumn{2}{c|}{MCD-23}      & \multicolumn{2}{c|}{MCD-21}                           & \multicolumn{2}{c}{MCD-23}       \\
                        & PSNR           & \multicolumn{1}{c|}{SSIM}            & PSNR           & SSIM            & PSNR           & \multicolumn{1}{c|}{SSIM}            & PSNR           & SSIM            \\ \hline
SwinIR~\cite{liang2021swinir}                  & 35.55          & \multicolumn{1}{c|}{0.9068}          & 35.63          & 0.9275          & 33.35          & \multicolumn{1}{c|}{0.8770}          & 33.25          & 0.8948          \\
HAT~\cite{chen2023activating}                     & 35.59          & \multicolumn{1}{c|}{0.9103}          & 35.74          & 0.9284          & 33.43          & \multicolumn{1}{c|}{0.8801}          & 33.24          & 0.8954          \\
SRFormer~\cite{zhou2023srformer}                & 35.54          & \multicolumn{1}{c|}{0.9078}          & 35.81          & 0.9284          & 33.19          & \multicolumn{1}{c|}{0.8788}          & 33.29          & 0.8953          \\
DAT~\cite{chen2023dual}                     & 36.09          & \multicolumn{1}{c|}{0.9144}          & 36.48          & 0.9323          & 33.73          & \multicolumn{1}{c|}{0.8841}          & 33.71          & 0.9025          \\
CFAT~\cite{ray2024cfat}                    & 35.62          & \multicolumn{1}{c|}{0.9105}          & 35.80          & 0.9287          & 33.34          & \multicolumn{1}{c|}{0.8800}          & 33.28          & 0.8960          \\
MambaIR~\cite{guo2024mambair}                 & 36.41          & \multicolumn{1}{c|}{0.9166}          & 36.73          & 0.9335          & 33.85          & \multicolumn{1}{c|}{0.8872}          & 34.03          & 0.9048          \\
ASSAN (Ours)                   & \textbf{36.66} & \multicolumn{1}{c|}{\textbf{0.9185}} & \textbf{37.33} & \textbf{0.9363} & \textbf{34.08} & \multicolumn{1}{c|}{\textbf{0.8911}} & \textbf{34.37} & \textbf{0.9080} \\ \hline
\end{tabular}
\end{center}
\label{table:result}
\end{table}

\noindent\textbf{Qualitative results.}
Following the common 3D vasculature visualization~\cite{leitgeb2014doppler,li2020deep,jiang2020comparative}, we present Maximum Intensity Projection (MIP) images for qualitative comparison in Fig.~\ref{fig:result}, comparing the MIP images of the volumes in MCD-23 $\times4$ sparsity. The primary evaluation factor of sparse ODT reconstruction is the reconstructed vessel structures. Fig.~\ref{fig:result} shows that, in most challenging cases, only our ASSAN successfully reconstructs the vessels. These results demonstrate the advantage of the proposed ASSAN over the other compared methods.

\subsection{Ablation Study}
\noindent\textbf{Scanning methods.}
Table~\ref{table:ablation} (left) presents results using different scanning methods on $\times 4$ sparsity.
SS2D (Fig.~\ref{fig:blockscan}(f)) refers to the method in MambaIR~\cite{guo2024mambair}. A-SS$\rightarrow$B-SS indicates that B-line Selective Scan (B-SS) is stacked after A-SS. A-SS$\rightarrow$B-SA applies the A-SS Block followed by B-SA, while A-SS $\rightarrow$ B-GA applies the A-SS Block followed by B-GA Block with the gated design. The results validate the design of our scanning strategy.

\begin{table}[!tbp]
\caption{Ablation study on scanning methods and feedforward networks.}
\begin{center}
\small
\begin{tabular}{lcccc|lcccc}
\hline
\multicolumn{5}{c|}{Comparison of scanning methods}                                                                                     & \multicolumn{5}{c}{Comparison of feedforward networks}                                                                                  \\ \hline
\multicolumn{1}{l|}{\multirow{2}{*}{Method}} & \multicolumn{2}{c|}{MCD-21}                           & \multicolumn{2}{c|}{MCD-23}      & \multicolumn{1}{l|}{\multirow{2}{*}{Method}} & \multicolumn{2}{c|}{MCD-21}                           & \multicolumn{2}{c}{MCD-23}       \\
\multicolumn{1}{l|}{}                        & PSNR           & \multicolumn{1}{c|}{SSIM}            & PSNR           & SSIM            & \multicolumn{1}{l|}{}                        & PSNR           & \multicolumn{1}{c|}{SSIM}            & PSNR           & SSIM            \\ \hline
\multicolumn{1}{l|}{SS2D}                    & 33.85          & \multicolumn{1}{c|}{0.8874}          & 33.94          & 0.9051          & \multicolumn{1}{l|}{w/o FFN}                 & 34.01          & \multicolumn{1}{c|}{0.8894}          & 34.29          & 0.9075          \\
\multicolumn{1}{l|}{A-SS $\to$ B-SS} & 33.89          & \multicolumn{1}{c|}{0.8883}          & 34.13          & 0.9063          & \multicolumn{1}{l|}{MLP}                     & 33.99          & \multicolumn{1}{c|}{0.8880}          & 34.28          & 0.9073          \\
\multicolumn{1}{l|}{A-SS $\to$ B-SA} & 33.92          & \multicolumn{1}{c|}{0.8884}          & 34.25          & 0.9068          & \multicolumn{1}{l|}{CAB}                     & 34.03          & \multicolumn{1}{c|}{0.8895}          & 34.28          & 0.9076          \\
\multicolumn{1}{l|}{B-GA $\to$ A-SS} & 33.98          & \multicolumn{1}{c|}{0.8890}          & 34.25          & 0.9074          & \multicolumn{1}{l|}{LEFN2D}                  & 34.03          & \multicolumn{1}{c|}{0.8895}          & 34.30          & 0.9073          \\
\multicolumn{1}{l|}{A-SS $\to$ B-GA} & \textbf{34.01} & \multicolumn{1}{c|}{\textbf{0.8894}} & \textbf{34.29} & \textbf{0.9075} & \multicolumn{1}{l|}{LEFN}                    & \textbf{34.08} & \multicolumn{1}{c|}{\textbf{0.8911}} & \textbf{34.37} & \textbf{0.9080} \\ \hline
\end{tabular}
\end{center}
\label{table:ablation}
\end{table}

\noindent\textbf{Feedforward Network.}
The results for different FFNs are shown in Table~\ref{table:ablation} (right). We compare the proposed LEFN with MLP~\cite{liang2021swinir}, CAB~\cite{zhang2018image,guo2024mambair}, and LEFN2D ($5\times5$ Conv2D). As MLP only performs channel mixing and CAB focuses on global information via channel attention, neither works well. LEFN2D also has limited impact due to its small receptive field and lack of explicit separation between A-line and B-line modeling. In contrast, the proposed LEFN effectively improves performance even with fewer parameters than LEFN2D.

\section{Conclusion}
We develop a novel ODT sparse reconstruction pipeline, named Alternative State Space Attention Network (ASSAN), to reduce scanning time and storage demands. To effectively model the unique information distributions along different axes in ODT B-scans, ASSAN processes each A-line through a state space model, followed by a gated self-attention block for modeling each B-line. A sequential Conv1D-based feedforward network is used to enhance local features. Extensive experiments indicate the effectiveness of the proposed pipeline.

\newpage
\bibliographystyle{splncs04}
\bibliography{ref.bib}

\begin{thebibliography}{10}
\providecommand{\url}[1]{\texttt{#1}}
\providecommand{\urlprefix}{URL }
\providecommand{\doi}[1]{https://doi.org/#1}

\bibitem{chen2021pre}
Chen, H., Wang, Y., Guo, T., Xu, C., Deng, Y., Liu, Z., Ma, S., Xu, C., Xu, C., Gao, W.: Pre-trained image processing transformer. In: CVPR. pp. 12299--12310 (2021)

\bibitem{chen2023activating}
Chen, X., Wang, X., Zhou, J., Qiao, Y., Dong, C.: Activating more pixels in image super-resolution transformer. In: CVPR. pp. 22367--22377 (2023)

\bibitem{chen2023dual}
Chen, Z., Zhang, Y., Gu, J., Kong, L., Yang, X., Yu, F.: Dual aggregation transformer for image super-resolution. In: ICCV. pp. 12312--12321 (2023)

\bibitem{chen2008doppler}
Chen, Z., Zhang, J.: Doppler optical coherence tomography. Optical Coherence Tomography: Technology and Applications pp. 621--651 (2008)

\bibitem{dai2019second}
Dai, T., Cai, J., Zhang, Y., Xia, S.T., Zhang, L.: Second-order attention network for single image super-resolution. In: CVPR. pp. 11065--11074 (2019)

\bibitem{elfwing2018sigmoid}
Elfwing, S., Uchibe, E., Doya, K.: Sigmoid-weighted linear units for neural network function approximation in reinforcement learning. Neural networks  \textbf{107},  3--11 (2018)

\bibitem{fang2016segmentation}
Fang, L., Li, S., Cunefare, D., Farsiu, S.: Segmentation based sparse reconstruction of optical coherence tomography images. TMI  \textbf{36}(2),  407--421 (2016)

\bibitem{gu2023mamba}
Gu, A., Dao, T.: Mamba: Linear-time sequence modeling with selective state spaces. arXiv preprint arXiv:2312.00752  (2023)

\bibitem{gu2021efficiently}
Gu, A., Goel, K., Re, C.: Efficiently modeling long sequences with structured state spaces. In: ICLR (2021)

\bibitem{gu2021combining}
Gu, A., Johnson, I., Goel, K., Saab, K., Dao, T., Rudra, A., R{\'e}, C.: Combining recurrent, convolutional, and continuous-time models with linear state space layers. NIPS  \textbf{34},  572--585 (2021)

\bibitem{guo2024mambair}
Guo, H., Li, J., Dai, T., Ouyang, Z., Ren, X., Xia, S.T.: Mambair: A simple baseline for image restoration with state-space model. In: ECCV (2024)

\bibitem{han2025demystify}
Han, D., Wang, Z., Xia, Z., Han, Y., Pu, Y., Ge, C., Song, J., Song, S., Zheng, B., Huang, G.: Demystify mamba in vision: A linear attention perspective. NIPS  \textbf{37},  127181--127203 (2025)

\bibitem{hendrycks2016gaussian}
Hendrycks, D., Gimpel, K.: Gaussian error linear units (gelus). arXiv preprint arXiv:1606.08415  (2016)

\bibitem{jiang2020comparative}
Jiang, Z., Huang, Z., Qiu, B., Meng, X., You, Y., Liu, X., Liu, G., Zhou, C., Yang, K., Maier, A., et~al.: Comparative study of deep learning models for optical coherence tomography angiography. Biomedical optics express  \textbf{11}(3),  1580--1597 (2020)

\bibitem{kalman1960new}
Kalman, R.E.: A new approach to linear filtering and prediction problems  (1960)

\bibitem{leitgeb2014doppler}
Leitgeb, R.A., Werkmeister, R.M., Blatter, C., Schmetterer, L.: Doppler optical coherence tomography. Progress in retinal and eye research  \textbf{41},  26--43 (2014)

\bibitem{li2020deep}
Li, A., Du, C., Volkow, N.D., Pan, Y.: A deep-learning-based approach for noise reduction in high-speed optical coherence doppler tomography. Journal of biophotonics  \textbf{13}(10),  e202000084 (2020)

\bibitem{li2023efficient}
Li, Y., Fan, Y., Xiang, X., Demandolx, D., Ranjan, R., Timofte, R., Van~Gool, L.: Efficient and explicit modelling of image hierarchies for image restoration. In: CVPR. pp. 18278--18289 (2023)

\bibitem{liang2021swinir}
Liang, J., Cao, J., Sun, G., Zhang, K., Van~Gool, L., Timofte, R.: Swinir: Image restoration using swin transformer. In: ICCV. pp. 1833--1844 (2021)

\bibitem{lim2017enhanced}
Lim, B., Son, S., Kim, H., Nah, S., Mu~Lee, K.: Enhanced deep residual networks for single image super-resolution. In: CVPR Workshops. pp. 136--144 (2017)

\bibitem{liu2024vmamba}
Liu, Y., Tian, Y., Zhao, Y., Yu, H., Xie, L., Wang, Y., Ye, Q., Liu, Y.: Vmamba: Visual state space model. arXiv preprint arXiv:2401.10166  (2024)

\bibitem{loshchilov2016sgdr}
Loshchilov, I., Hutter, F.: Sgdr: Stochastic gradient descent with warm restarts. In: ICLR (2016)

\bibitem{marques2018review}
Marques, E.C., Maciel, N., Naviner, L., Cai, H., Yang, J.: A review of sparse recovery algorithms. IEEE access  \textbf{7},  1300--1322 (2018)

\bibitem{ning2015sparse}
Ning, L., Laun, F., Gur, Y., DiBella, E.V., Deslauriers-Gauthier, S., Megherbi, T., Ghosh, A., Zucchelli, M., Menegaz, G., Fick, R., et~al.: Sparse reconstruction challenge for diffusion mri: Validation on a physical phantom to determine which acquisition scheme and analysis method to use? Medical image analysis  \textbf{26}(1),  316--331 (2015)

\bibitem{niu2020single}
Niu, B., Wen, W., Ren, W., Zhang, X., Yang, L., Wang, S., Zhang, K., Cao, X., Shen, H.: Single image super-resolution via a holistic attention network. In: ECCV. pp. 191--207. Springer (2020)

\bibitem{potter2008sparse}
Potter, L.C., Schniter, P., Ziniel, J.: Sparse reconstruction for radar. In: Algorithms for Synthetic Aperture Radar Imagery XV. vol.~6970, pp. 9--23. SPIE (2008)

\bibitem{ray2024cfat}
Ray, A., Kumar, G., Kolekar, M.H.: Cfat: Unleashing triangular windows for image super-resolution. In: CVPR. pp. 26120--26129 (2024)

\bibitem{shi2016real}
Shi, W., Caballero, J., Husz{\'a}r, F., Totz, J., Aitken, A.P., Bishop, R., Rueckert, D., Wang, Z.: Real-time single image and video super-resolution using an efficient sub-pixel convolutional neural network. In: CVPR. pp. 1874--1883 (2016)

\bibitem{spaide2018optical}
Spaide, R.F., Fujimoto, J.G., Waheed, N.K., Sadda, S.R., Staurenghi, G.: Optical coherence tomography angiography. Progress in retinal and eye research  \textbf{64},  1--55 (2018)

\bibitem{szegedy2016rethinking}
Szegedy, C., Vanhoucke, V., Ioffe, S., Shlens, J., Wojna, Z.: Rethinking the inception architecture for computer vision. In: CVPR. pp. 2818--2826 (2016)

\bibitem{vaswani2017attention}
Vaswani, A., Shazeer, N., Parmar, N., Uszkoreit, J., Jones, L., Gomez, A.N., Kaiser, {\L}., Polosukhin, I.: Attention is all you need. NIPS  \textbf{30} (2017)

\bibitem{wang2007three}
Wang, R.K., Jacques, S.L., Ma, Z., Hurst, S., Hanson, S.R., Gruber, A.: Three dimensional optical angiography. Optics express  \textbf{15}(7),  4083--4097 (2007)

\bibitem{wei2010sparse}
Wei, S.J., Zhang, X.L., Shi, J., Xiang, G.: Sparse reconstruction for sar imaging based on compressed sensing. Progress in electromagnetics research  \textbf{109},  63--81 (2010)

\bibitem{yang2016sparse}
Yang, A.C., Kretzler, M., Sudarski, S., Gulani, V., Seiberlich, N.: Sparse reconstruction techniques in magnetic resonance imaging: methods, applications, and challenges to clinical adoption. Investigative radiology  \textbf{51}(6),  349--364 (2016)

\bibitem{yang2024gated}
Yang, S., Wang, B., Shen, Y., Panda, R., Kim, Y.: Gated linear attention transformers with hardware-efficient training. In: ICML (2024)

\bibitem{zhang2024transcending}
Zhang, L., Li, Y., Zhou, X., Zhao, X., Gu, S.: Transcending the limit of local window: Advanced super-resolution transformer with adaptive token dictionary. In: CVPR. pp. 2856--2865 (2024)

\bibitem{zhang2018image}
Zhang, Y., Li, K., Li, K., Wang, L., Zhong, B., Fu, Y.: Image super-resolution using very deep residual channel attention networks. In: ECCV. pp. 286--301 (2018)

\bibitem{zhang2015survey}
Zhang, Z., Xu, Y., Yang, J., Li, X., Zhang, D.: A survey of sparse representation: algorithms and applications. IEEE access  \textbf{3},  490--530 (2015)

\bibitem{zhao2000phase}
Zhao, Y., Chen, Z., Saxer, C., Xiang, S., de~Boer, J.F., Nelson, J.S.: Phase-resolved optical coherence tomography and optical doppler tomography for imaging blood flow in human skin with fast scanning speed and high velocity sensitivity. Optics letters  \textbf{25}(2),  114--116 (2000)

\bibitem{zhou2023srformer}
Zhou, Y., Li, Z., Guo, C.L., Bai, S., Cheng, M.M., Hou, Q.: Srformer: Permuted self-attention for single image super-resolution. In: ICCV. pp. 12780--12791 (2023)

\end{thebibliography}

\end{document}